# Deep Network Ensemble Learning applied to Image Classification using CNN Trees

Abdul Mueed Hafiz*[1] and Ghulam Mohiuddin Bhat[2]

[1, 2] Department of Electronics and Communication Engineering
Institute of Technology, University of Kashmir
Srinagar, J&K, India, 190006.

*[1]Corresponding Author Email: mueedhafiz@uok.edu.in
[2]Co-author Email: drgmbhat@uok.edu.in

ORC-ID[1]: 0000-0002-2266-3708
ORC-ID[2]: 0000-0001-9106-4699

*Abstract* - Traditional machine learning approaches may fail to perform satisfactorily when dealing with complex data. In this context, the importance of data mining evolves w.r.t. building an efficient knowledge discovery and mining framework. Ensemble learning is aimed at integration of fusion, modeling and mining of data into a unified model. However, traditional ensemble learning methods are complex and have optimization or tuning problems. In this paper, we propose a simple, sequential, efficient, ensemble learning approach using multiple deep networks. The deep network used in the ensembles is ResNet50. The model draws inspiration from binary decision/classification trees. The proposed approach is compared against the baseline viz. the single classifier approach i.e. using a single multiclass ResNet50 on the *ImageNet* and *Natural Images* datasets. Our approach outperforms the baseline on all experiments on the *ImageNet* dataset. Code is available in *https://github.com/mueedhafiz1982/CNNTreeEnsemble*

## 1. Introduction

In spite of significant successes achieved in the field of knowledge discovery, traditional machine learning techniques can fail to perform satisfactorily when dealing with complex data, e.g. if it is imbalanced, high-dimensional, noisy, etc. This is mainly because of the fact that it is difficult for these techniques to capture various characteristics and underlying data structure [1]. In this context, an important topic evolves in the field of data mining as how to effectively build an efficient model for knowledge discovery and mining. Ensemble learning, which is one research hotspot, is aimed at integration of fusion, modeling, and mining of data into a unified model. Specifically, ensemble learning first extracts feature sets with various transformations. Based on the learnt features, several learning algorithms produce weak prediction results. In the end, ensemble learning performs fusion of the information obtained from the above for knowledge discovery and higher prediction performance with voting schemes in an adaptive manner [1].

A decent introduction to ensemble learning can be found in the recent survey of Dong et. al. [1]. In case of transfer learning based ensembles [1], related works like [2] have used ensembles for combination of outputs of different layer-based transfer conditions of deep networks. Experimentation has shown that it reduces the effects of adverse feature transference of features on tasks related to image recognition [1]. Nozza et al. [3] have also used ensembles to reduce the cross-domain error for the problem of domain adaptation for the task of sentiment classification. Liu et al. [4] have designed an transfer-learning based ensemble framework which uses AdaBoost [5] for weight adjustment of source data and target data. The technique achieved decent performance using UCI datasets in case of insufficient data. More work [6-14] has also been done.

In [9], the authors state that neural networks lag behind the state-of-the-art algorithms for Time Series Classification (TSC). The latter are composed of ensembles of 37 classifiers which are non deep-learning based. The authors attribute this lag to the lack of deep network ensembles for TSC. In their work, they show that how an ensemble of 60 deep networks can significantly improve the state-of-the-art performance of deep networks for TSC. They use the UCR/UEA archive [15], which is the largest publicly available database for time series analysis. Our paper is about image recognition and does not go in the direction of this research i.e. TSC.

In [10], the authors state that contemporary deep networks suffer from problems like difficulty of interpretation, and overfitting. Although regularization techniques have been investigated to avoid overfitting, but without too much underlying theoretical framework. The author argues that in order to extract neural network features for decision making, it is important to consider the cluster paths in neural networks. These features are particularly of interest because they give the actual decision making process in the neural network. The author accordingly presents an ensemble technique for neural networks which guarantees test accuracy. The ensemble technique has given state-of-the-art results for ResNets [16-18] on CIFAR-10 [19] and has also improved performance of various models applied to it. However, CIFAR-10 images are too small and their relevance is lesser.

In [6], the authors state that the sharing of medical image datasets among various institutions is limited by legal issues. As a result, medical research that requires large datasets suffers considerably. This is because modern machine learning techniques commonly require immensely large imaging datasets. They introduce constrained Generative Adversarial Network ensembles (cGANe) in order to address this problem by altering the imaging data with information preservation, enabling the reproduction of research elsewhere with the shareable data. However, the authors state that the applicability of the proposed technique needs further validation with a range of medical image data types.

In [20], the authors state that recent evidence has revealed that Neural Machine Translation (NMT) models with deep neural networks may be more effective. However, they are difficult to train. The authors present a MultiScale Collaborative (MSC) framework in order to ease NMT model training. The models used are much deeper than the previous ones. They give evidence demonstrating that the MSC networks are easily optimized and can give quality improvements by considerably increasing depth. On translation tasks with three translation directions, their extremely deep model surpasses significantly outperforms the state-of-the-art deep NMT models. However, using very deep networks is fraught with training difficulty and parameter overload.

As is also observed from above, ensemble learning is complex, and has optimization issues. We propose a simple, efficient, sequential, two-layer ensemble learning framework, which is similar to decision/classification trees in some aspects. It has the advantage of outperforming the baseline standalone deep models like ResNet50 [10, 16, 21] on datasets like *ImageNet* [22-24]. On the *Natural Images* [25] dataset, the classification count of our approach lags behind that of the competitor, by only one test image. The ensemble models themselves are made of no-top ResNets with augmented end layers and transfer learning is used on the augmented end-layers, hence taking lesser time than training the full ResNet. It should be noted that the baseline is not extended to hybrid classifiers which use CNN feature maps, because these hybrid classifiers ca be in turn used with the CNNs in the proposed ensembles. Also, we did not report low classification accuracy yielding techniques which use images as input.

## 2. Background

An ensemble by itself is a supervised learning algorithm, because it can be trained and subsequently used for making predictions. A trained ensemble, hence, represents one hypothesis. However, this hypothesis is not necessarily contained inside the hypothesis space of various models which build it. Hence, ensembles may be shown to have higher flexibility with regards to the functions that they represent. Empirically, ensembles yield better results for the case when diversity is significant among the models. [26, 27] Most ensemble techniques, hence, promote diversity among their models. [28, 29] By using strong learning algorithms, better performance has been obtained as compared to using algorithms that dumb-down the ensemble models for promoting diversity. [30]

Prediction by an ensemble typically is computationally more expensive than that of a single model, hence ensembles in a way compensate for single model learning algorithms by performing more computation. Fast techniques such as decision trees are commonly used as ensembles, although slower techniques may benefit from ensemble learning as well. By analogy, ensemble learning has also been used in unsupervised learning, e.g. in consensus clustering, anomaly detection, etc.

Our technique borrows concepts from decision trees as it also splits the data into classes and refines the number of classes as it goes deeper down the tree,. However, it cannot be directly compared to decision trees as it does not involve feature splitting. It mostly does binary classification on all the classes, and then monitors the final prediction values (softmax or sigmoid) in a decision chain (*if...else*) to make a decision on the class of the test sample. We do not use voting hence our technique is different from conventional ensemble approaches. The advantages of our approach are simplicity of the overall framework and higher recognition accuracy as compared to the baseline drawn from the normal scenario i.e. when an ensemble is not used, vis-à-vis a multiclass Convolutional Neural Network (CNN), in this case ResNet50 [16-18]. The ensemble classifiers are ResNets themselves which are the no-top version, augmented with end layers. The training using transfer learning is done on the augmented layers at end of the no-top ResNet50. The datasets used for performance evaluation are *ImageNet* [22-24] and *Natural Images* [25]. Figure 1 shows the overview of our technique.

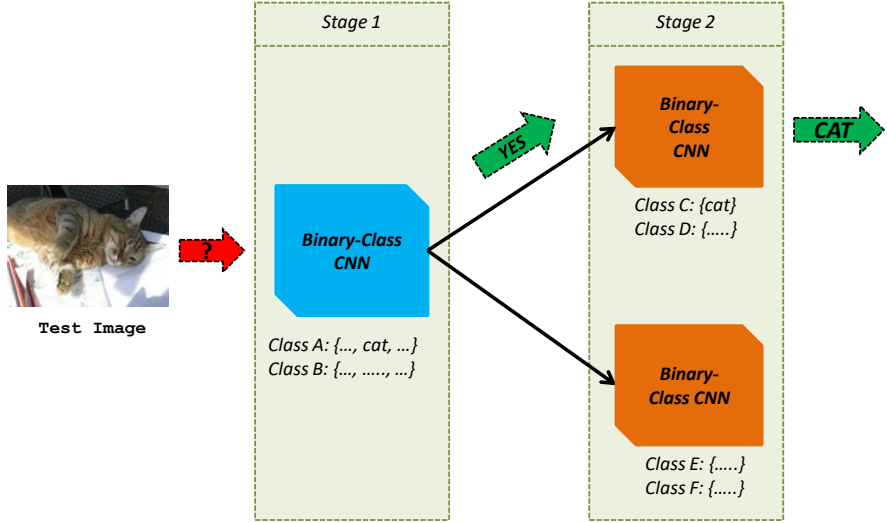

**Figure 1.** Overview of proposed technique

## 3. Implementation and Experimentation

The hardware used for experimentation comprises of an Intel® Xeon® processor (with 2 cores), 16 GB RAM and 12 GB GPU. We have used Tensorflow to implement the algorithm and the CNNs in Python. The CNNs used is ResNet50 [17]. The datasets used are ImageNet [22-24] and Natural Images [25]. The input to the CNNs is set to (*224, 224, 3*). All the CNNs use *Adam* Optimizer for training, with mini-batch size of *32* for larger image categories and of *16* for smaller ones, respectively. The CNNs are trained using early-stopping by monitoring *validation loss* with *patience* = 7. The loss function used is *binary cross-entropy* for binary-class CNNs and *categorical cross-entropy* for multiclass CNNs. The predicted scores of the final layer of the CNNs are used for decision making. We divide the experiments into two sets based on the number of classes used for classification, i.e. four and six, respectively.

### 3.1 Four-Class Classification

We first focus on four class image recognition, using the conventional multiclass CNNs and the proposed technique on the datasets. The network architecture of conventional multiclass CNN used is shown in Figure 1.

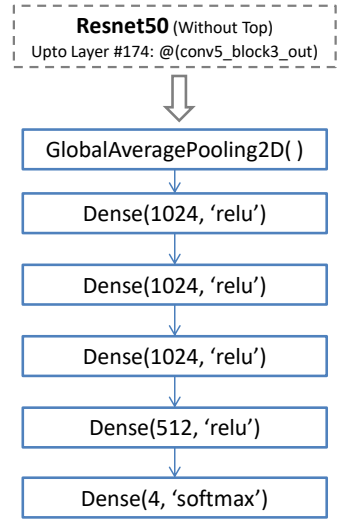

**Figure 1.** Network Architectures used for conventional multiclass CNN implementation (#classes=4).

Let the four classes be denoted by *c1, c2, c3* and *c4*, respectively. The four-class ensemble uses a single binary-class CNN (2 categories per class) i.e. [ (*c1, c2*) v/s (*c3, c4*) ] in first stage, and two binary-class CNNs in second stage i.e. [ *c1* v/s *c2* ] and [ *c3* v/s *c4* ]. Figure 2 shows the architecture of the four-class classification ensemble for the proposed technique.

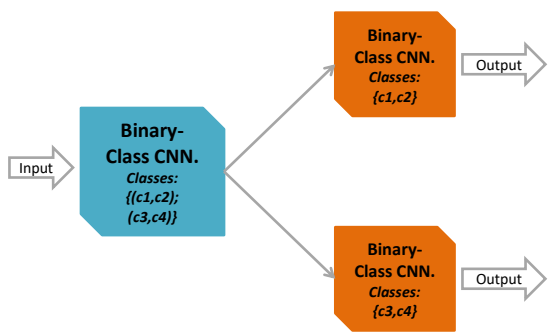

**Figure 2**. Ensemble Structure for Four-Class Classification using Proposed Technique.

The architecture of the binary-class CNNs used in the ensemble is same as given in Figure 1, except that their respective final layers have 2 neurons each. Figure 3 shows the individual CNN architecture for the binary-class CNNs used in ensemble shown in Figure 2.

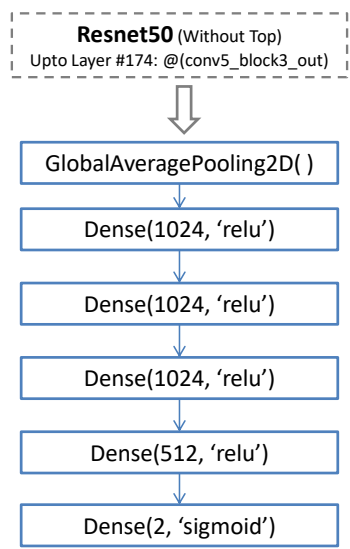

**Figure 3.** Network Architecture of binary-class CNNs used in four-class ensemble[17]

The details of the image subsets used here are given in Table 1.

**Table 1**. Distribution of experimental data

| Dataset | Classes Used (c1, c2, c3, c4) | Training Images | Validation Images | Testing Images |
|---|---|---|---|---|
| ImageNet [24] | 4 (*Bikes, Ships, Tractors, Wagons*) | 1531 | 788 | 745 |
| Natural Images [25] | 4 (*Airplane, Motorbike, Car, Person*) | 1515 | 800 | 800 |

The details of the experimentation for four-class classification are shown in Table 2.

**Table 2.** Accuracy on various approaches on different datasets for four-class classification

| Approach | ResNet50 (4-class CNN) | Proposed Approach Using ResNet50 |
|---|---|---|
| **ImageNet [24]** | .9007 | **.9074** |
| **Natural Images [25]** | 1.0 | .9988 |

It should be noted that the test set classification count of the proposed approach for *Natural Images* subset lags behind that of its competitor (multiclass RseNet50) by only one image.

### 3.2 Six-Class Classification

Let the categories be *c1, c2, c3, c4, c5* and *c6*. The distribution of the images for this task in ImageNet [23] is given in Table 3.

**Table 3.** Distribution of images (*ImageNet* [24]) for six-class classification task

| Dataset | Classes Used (c1, c2, c3, c4, c5, c6) | Training Images | Validation Images | Testing Images |
|---|---|---|---|---|
| ImageNet [24] | 6 (*Bikes, Ships, Tractors, Wagons, Cats, Dogs*) | 2214 | 1073 | 1022 |

We have implemented six-class classification using two different ensemble structures. The CNN architecture used in this ensemble is akin to that used in earlier one except that here the final layer may have 6, 3 or 2 neurons for six-, three-, or binary-class CNN respectively. When final layer neuron count is 2, *sigmoid* activation is used. For more that 2 final layer neuron count, *softmax* activation is used. The experimentation on these two ensemble structures is discussed below.

### 3.2.1 Classification using Ensemble Structure #1

The first ensemble uses a three-class CNN (3 categories per class) i.e. [ ($c1$, $c2$) v/s ($c3$, $c4$) v/s ($c5$, $c6$) ], in first stage, and three binary-class CNNs in second stage i.e. [ $c1$ v/s $c2$ ], [ $c3$ v/s $c4$ ], and [ $c5$ v/s $c6$ ]. Figure 4 shows the architecture of the six-class classification ensemble#1 for the proposed technique.

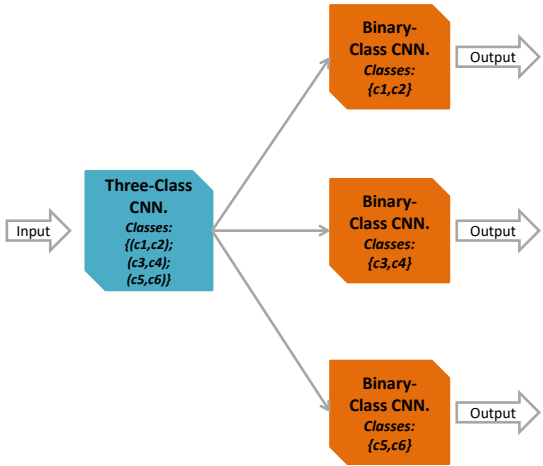

**Figure 4**. Ensemble Structure *#1* for Six-Class Classification using Proposed Technique.

The test images are resized to various sizes before classification by trained CNNs (in ensemble) for further investigation. Since each second-stage binary-CNN used in the proposed ensembles has its own specific response as per testing image size. The best-result yielding testing image size was found from testing the second stage binary CNNs on the validation image subsets. Accordingly, class-specific CNN image-size can be used for testing in our technique. This cannot be done for a conventional multiclass CNN. The results of the experiments are shown in Table 4.

**Table 4.** Accuracy on various datasets on ImageNet [24] for six-class classification using ensemble structure *#1*

| Test Image Size | ResNet50 (6-class CNN) | Proposed Approach Using ResNet50 |
|---|---|---|
| *(224,224)* | .8963 | .8904 |
| *(299,299)* | .8982 | .8962 |
| CNN-Specific: (224,224) or (299,299) per CNN | - | .9061 |

### 3.2.2 Classification using Ensemble Structure #2

The second ensemble uses a single binary-class CNN (2 categories per class) i.e. [ ($c1$, $c2$, $c3$) v/s ($c5$, $c6$, $c4$) ], in first stage, and two three-class CNNs in second stage i.e. [ $c1$ v/s $c2$ v/s $c3$ ] and [ $c5$ v/s $c6$ v/s $c4$ ]. Figure 5 shows the architecture of the six-class classification ensemble#2 for the proposed technique.

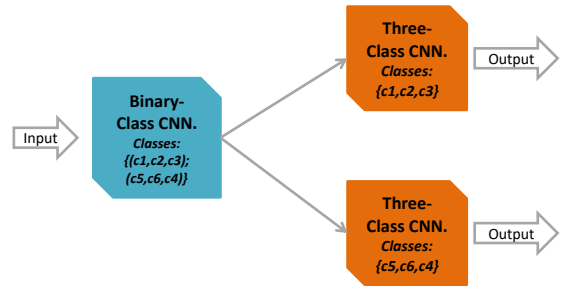

**Figure 5**. Ensemble Structure *#2* for Six-Class Classification using Proposed Technique.

The results of the experiments are shown in Table 5. No test image resizing was done in this case.

**Table 5.** Accuracy on various datasets on ImageNet [24] for six-class classification using ensemble structure *#2*

| Test Image Size | ResNet50 (4-class CNN) | Proposed Approach Using ResNet50 |
|---|---|---|
| *(224,224)* | .8963 | **.9002** |

It is observed that the classification accuracies of the proposed approach are higher than that of the baseline in all experiments using ImageNet subsets. The accuracy is also almost equal for the Natural Image subset.

## 4. Conclusion

The importance of using simple, efficient, easy to optimize, and successfully explored frameworks, for ensemble learning has been highlighted. Accordingly, a new deep network based ensemble learning approach has been proposed. The proposed approach is inspired by decision/classification trees in some aspects, though its working is different. Various models are implemented, and improvement in performance w.r.t. baseline is observed. The baseline comprises of the non-ensemble approach, viz. a multiclass standalone CNN, i.e. ResNet50. It should be noted that the baseline is chosen for comparison because each model in the ensemble itself is a ResNet50 CNN. The experimentation demonstrates that our approach outperforms the baseline on all models for the *ImageNet* dataset. On the *Natural Images* dataset, the classification count of our approach lags behind that of its competitor by only one image. Other hybrid techniques can be applied to both the proposed approach as well as the baseline i.e. multiclass ResNet50 classifier, and are not used. Low classification accuracy yielding techniques for images are not reported. Future work would be based on these points.

## Conflict of Interest

The authors declare no conflict of interest.

## References

[1] X. Dong, Z. Yu, W. Cao, Y. Shi, and Q. Ma, "A survey on ensemble learning," *Frontiers of Computer Science,* vol. 14, no. 2, pp. 241-258, 2020/04/01 2020, doi: 10.1007/s11704-019-8208-z.

[2] C. Kandaswamy, L. M. Silva, L. A. Alexandre, and J. M. Santos, "Deep Transfer Learning Ensemble for Classification," Cham, 2015: Springer International Publishing, in Advances in Computational Intelligence, pp. 335-348.

[3] D. Nozza, E. Fersini, and E. Messina, "Deep Learning and Ensemble Methods for Domain Adaptation," in *2016 IEEE 28th International Conference on Tools with Artificial Intelligence (ICTAI)*, 6-8 Nov. 2016 2016, pp. 184-189, doi: 10.1109/ICTAI.2016.0037.

[4] X. Liu, Z. Liu, G. Wang, Z. Cai, and H. Zhang, "Ensemble Transfer Learning Algorithm," *IEEE Access,* vol. 6, pp. 2389-2396, 2018, doi: 10.1109/ACCESS.2017.2782884.

[5] Y. Freund, R. Schapire, and N. Abe, "A short introduction to boosting," *Journal-Japanese Society For Artificial Intelligence,* vol. 14, no. 771-780, p. 1612, 1999.

[6] E. Dikici, L. M. Prevedello, M. Bigelow, R. D. White, and B. S. Erdal, "Constrained Generative Adversarial Network Ensembles for Sharable Synthetic Data Generation," *arXiv preprint arXiv:2003.00086,* 2020.

[7] Z. Yu *et al.*, "Multiobjective Semisupervised Classifier Ensemble," *IEEE Transactions on Cybernetics,* vol. 49, no. 6, pp. 2280-2293, 2019, doi: 10.1109/TCYB.2018.2824299.

[8] Z. Yu *et al.*, "Adaptive Semi-Supervised Classifier Ensemble for High Dimensional Data Classification," *IEEE Transactions on Cybernetics,* vol. 49, no. 2, pp. 366-379, 2019, doi: 10.1109/TCYB.2017.2761908.

[9] H. I. Fawaz, G. Forestier, J. Weber, L. Idoumghar, and P. Muller, "Deep Neural Network Ensembles for Time Series Classification," in *2019 International Joint Conference on Neural Networks (IJCNN)*,

14-19 July 2019 2019, pp. 1-6, doi: 10.1109/IJCNN.2019.8852316.
[10] S. Tao, "Deep Neural Network Ensembles," Cham, 2019: Springer International Publishing, in Machine Learning, Optimization, and Data Science, pp. 1-12.
[11] S. Sun, S. Wang, Y. Wei, and G. Zhang, "A Clustering-Based Nonlinear Ensemble Approach for Exchange Rates Forecasting," *IEEE Transactions on Systems, Man, and Cybernetics: Systems,* 2018.
[12] O. Sagi and L. Rokach, "Ensemble learning: A survey," *WIREs Data Mining and Knowledge Discovery,* vol. 8, no. 4, p. e1249, 2018, doi: 10.1002/widm.1249.
[13] K. Yang *et al.*, "Hybrid Classifier Ensemble for Imbalanced Data," *IEEE Transactions on Neural Networks and Learning Systems,* vol. 31, no. 4, pp. 1387-1400, 2020, doi: 10.1109/TNNLS.2019.2920246.
[14] J. Zheng, X. Cao, B. Zhang, X. Zhen, and X. Su, "Deep Ensemble Machine for Video Classification," *IEEE Transactions on Neural Networks and Learning Systems,* vol. 30, no. 2, pp. 553-565, 2019, doi: 10.1109/TNNLS.2018.2844464.
[15] Y. Chen *et al. The UCR time series classification archive*. [Online]. Available: www.cs.ucr.edu/~eamonn/time_series_data/
[16] S. Xie, R. Girshick, P. Dollár, Z. Tu, and K. He, "Aggregated Residual Transformations for Deep Neural Networks," in *2017 IEEE Conference on Computer Vision and Pattern Recognition (CVPR)*, 21-26 July 2017 2017, pp. 5987-5995, doi: 10.1109/CVPR.2017.634.
[17] K. He, X. Zhang, S. Ren, and J. Sun, "Deep residual learning for image recognition," in *Proceedings of the IEEE conference on computer vision and pattern recognition*, 2016, pp. 770-778.
[18] K. He, X. Zhang, S. Ren, and J. Sun, "Identity mappings in deep residual networks," in *European conference on computer vision*, 2016: Springer, pp. 630-645.
[19] A. Krizhevsky and G. Hinton, "Learning multiple layers of features from tiny images," 2009.
[20] X. Wei, H. Yu, Y. Hu, Y. Zhang, R. Weng, and W. Luo, "Multiscale Collaborative Deep Models for Neural Machine Translation," *arXiv preprint arXiv:2004.14021,* 2020.
[21] J. Redmon and A. Farhadi, "YOLO9000: Better, Faster, Stronger," in *2017 IEEE Conference on Computer Vision and Pattern Recognition (CVPR)*, 21-26 July 2017 2017, pp. 6517-6525, doi: 10.1109/CVPR.2017.690.
[22] O. Russakovsky *et al.*, "ImageNet Large Scale Visual Recognition Challenge," *International Journal of Computer Vision,* vol. 115, no. 3, pp. 211-252, 2015/12/01 2015, doi: 10.1007/s11263-015-0816-y.
[23] A. Krizhevsky, I. Sutskever, and G. E. Hinton, "Imagenet classification with deep convolutional neural networks," in *Advances in neural information processing systems*, 2012, pp. 1097-1105.
[24] J. Deng, W. Dong, R. Socher, L. Li, L. Kai, and F.-F. Li, "ImageNet: A large-scale hierarchical image database," in *2009 IEEE Conference on Computer Vision and Pattern Recognition*, 20-25 June 2009 2009, pp. 248-255, doi: 10.1109/CVPR.2009.5206848.
[25] P. Roy, S. Ghosh, S. Bhattacharya, and U. Pal, "Effects of degradations on deep neural network architectures," *arXiv preprint arXiv:1807.10108,* 2018.
[26] L. I. Kuncheva and C. J. Whitaker, "Measures of Diversity in Classifier Ensembles and Their Relationship with the Ensemble Accuracy," *Machine Learning,* vol. 51, no. 2, pp. 181-207, 2003/05/01 2003, doi: 10.1023/A:1022859003006.
[27] P. Sollich and A. Krogh, "Learning with ensembles: how over-fitting can be useful," presented at the Proceedings of the 8th International Conference on Neural Information Processing Systems, Denver, Colorado, 1995.
[28] G. Brown, J. Wyatt, R. Harris, and X. Yao, "Diversity creation methods: a survey and categorisation," *Information Fusion,* vol. 6, no. 1, pp. 5-20, 2005/03/01/ 2005, doi: https://doi.org/10.1016/j.inffus.2004.04.004.
[29] J. J. G. Adeva, U. Beresi, and R. Calvo, "Accuracy and diversity in ensembles of text categorisers," *CLEI Electronic Journal,* vol. 9, no. 1, pp. 1-12, 2005.
[30] M. Gashler, C. Giraud-Carrier, and T. Martinez, "Decision Tree Ensemble: Small Heterogeneous Is Better Than Large Homogeneous," in *2008 Seventh International Conference on Machine Learning and Applications*, 11-13 Dec. 2008 2008, pp. 900-905, doi: 10.1109/ICMLA.2008.154.

**Biography**

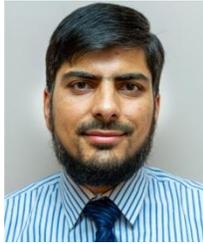

Abdul Mueed Hafiz completed his B.Tech. in Electronics & Communication Engineering, from Department of ECE, National Institute of Technology, Srinagar in 2005, his M.Tech. in Communication & Information Technology, from Department of ECE, NIT Srinagar in 2008, and his Ph.D. in Electronics (in the field of Artificial Intelligence), from PG Department of Electronics and Instrumentation Technology, University of Kashmir in 2018. Currently he works as an Assistant Professor (on substantive basis) in the Department of ECE., Institute of Technology, University of Kashmir, Srinagar, J&K, India. His research interests include computer vision, neural networks and learning systems. He has 14 publications in international journals / book chapters / international conferences. He is a reviewer for various journals like *IEEE TNNLS, Springer NEPL, ACM TALLIP*, etc. He is also a member of IEEE and ACM.

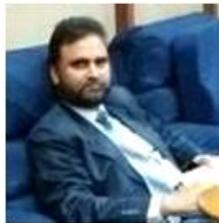

Ghulam Mohiuddin Bhat completed his M.Sc. in Electronics from University of Kashmir, Srinagar, J&K, India, his M.Tech. in Electronics & Communication Engineering from Z. H. College of Engineering & Technology, Aligarh Muslim University (AMU), Aligarh, India., and his Ph.D. in Electronics Engineering from Z. H. College of Engineering & Technology, Aligarh Muslim University (AMU), Aligarh, India. Currently, he serves as Head of Department and is a full-time Professor in Dept. of ECE, Institute of Technology, University of Kashmir, Srinagar, J&K, India. His research interests include Signal Processing and Automation. He has more than 200 publications in international journals / book chapters / international conferences. He has also served as Dean Engineering, University of Kashmir, Srinagar, and is former Director of his institute. He is also in charge of various national-level (Govt. Of India Sponsored) projects in J&K, India.